\documentclass{article}

\usepackage{times}
\usepackage{icip}
\usepackage{amsmath}
\usepackage{amssymb}
\usepackage{lipsum}
\usepackage{tabulary}
\usepackage{booktabs}
\usepackage{adjustbox}
\usepackage{subfloat}
\usepackage{float}
\usepackage{graphicx}
\usepackage{subcaption}

\title{
Attention Neural Network for Trash Detection on Water Channels 
}

\name{Mohbat Tharani, Abdul Wahab Amin, Mohammad Maaz and Murtaza Taj\thanks{This work was supported by Higher Education Commission, Pakistan under funding of National Agricultural Robotics Lab.}}
\address{Computer Vision and Graphics Lab, School of Science and Engineering, \\Lahore University of Management Sciences, Lahore, Pakistan}

\begin{document}

\maketitle

\thispagestyle{empty}
\pagestyle{empty}

\begin{abstract}
Rivers and canals flowing through cities are often used illegally for dumping trash. This contaminates fresh water channels as well as causes blockage in sewerage resulting urban flooding. When this contaminated water reaches agricultural fields, it results in degradation of soil and poses critical environmental as well as economic threats. The dumped trash is often found floating on the water surface. The trash could be disfigured, partially submerged, decomposed into smaller pieces, clumped together with other objects which obscures its shape and creates a challenging detection problem. This paper proposes a method for detection of visible trash floating on the water surface of the canals in urban areas. We also provide a large dataset, first of its kind, trash in water channels that contains object level annotations. A novel attention layer is proposed that improves the detection of smaller objects. Towards the end of this paper, we provide a detailed comparison of our method with state-of-the-art object detectors and show that our method significantly improves the detection of smaller objects. The dataset will be made publicly available.

\end{abstract}


\section{INTRODUCTION}
\label{introduction}
Every year millions of tons of trash, especially plastic, is discarded globally which pollutes our lands, rivers, and oceans. This causes environmental as well as economic repercussions. In developing countries, $90\%$ of sewerage and $70\%$ of the industrial waste is discharged in local water channels without treatment~\cite{liyanage2017impact} which contaminates water and adds toxins to our food chain. According to the United Nations world water development report~\cite{engin2018united}, annually about $3.5$ million people, mostly children, die from water-related infections. 

To cater to this issue of water pollution, the first step is to identify the main elements present in water. Trash is one of the major contributors which is dumped in drainage and fresh water channels of urban areas, from where it finally reaches the rivers. This trash consists of soluble and insoluble trash such as papers, card-boards, food residuals, plastic bottles and bags, etc. These trash elements upon reaching agricultural fields degrade the soil, reduce fertility and harm crops. To measure the amount of trash in canals as an index of water contamination, the detection of visual floating trash is a key step. The detected trash would then be quantified to notify planning authorities to take appropriate actions. 

\begin{figure}[t]
\centering
\begin{center}
\includegraphics [width=3.5cm, height=2.3cm]{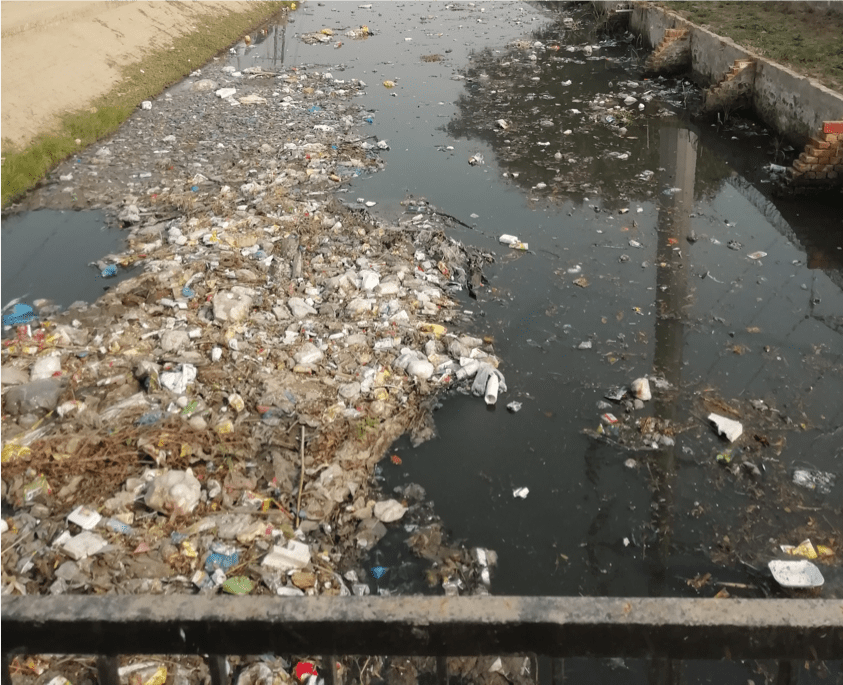} 
\includegraphics [width=3.5cm, height=2.3cm]{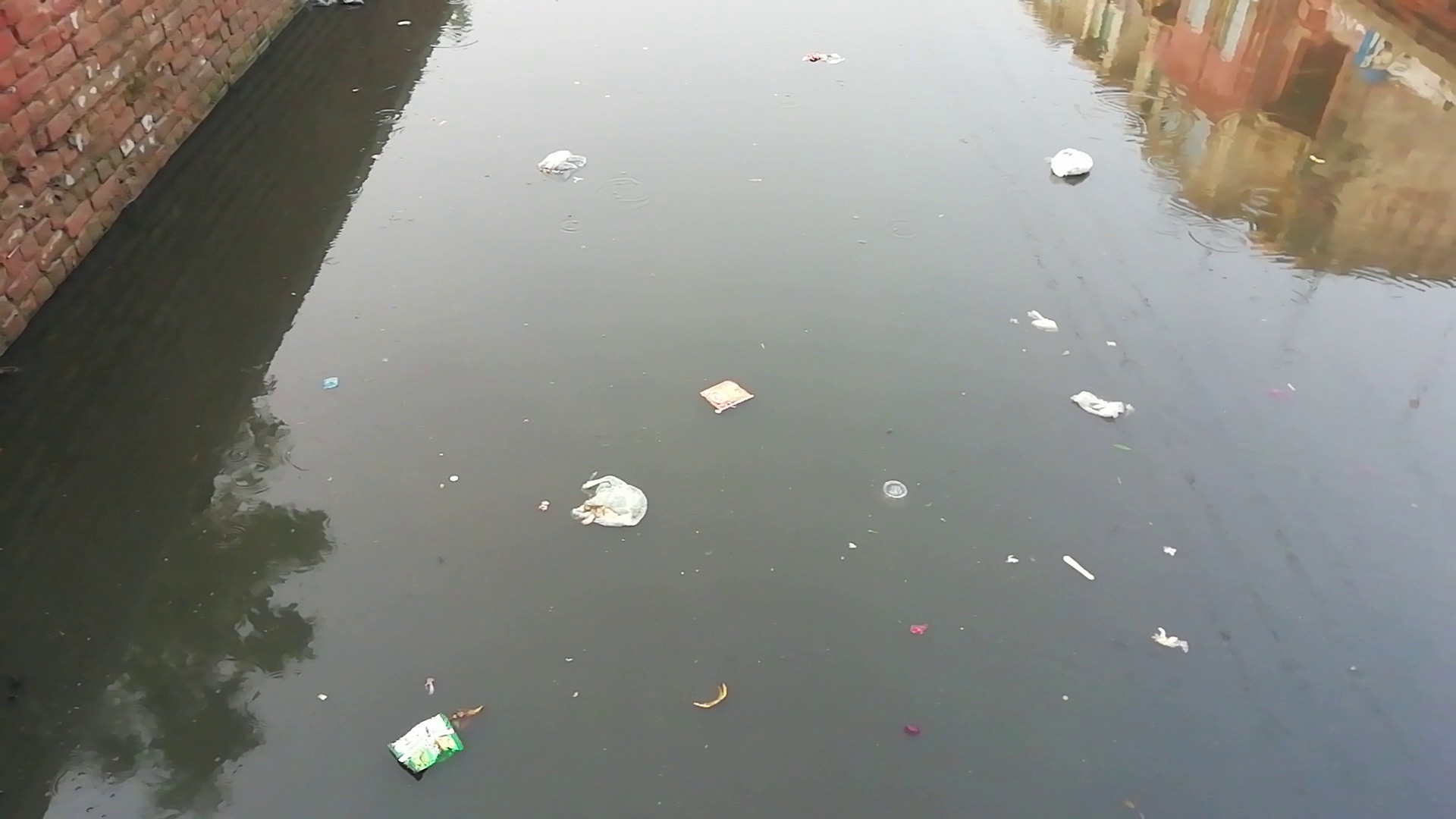}  \\
\includegraphics [width=3.5cm, height=2.3cm]{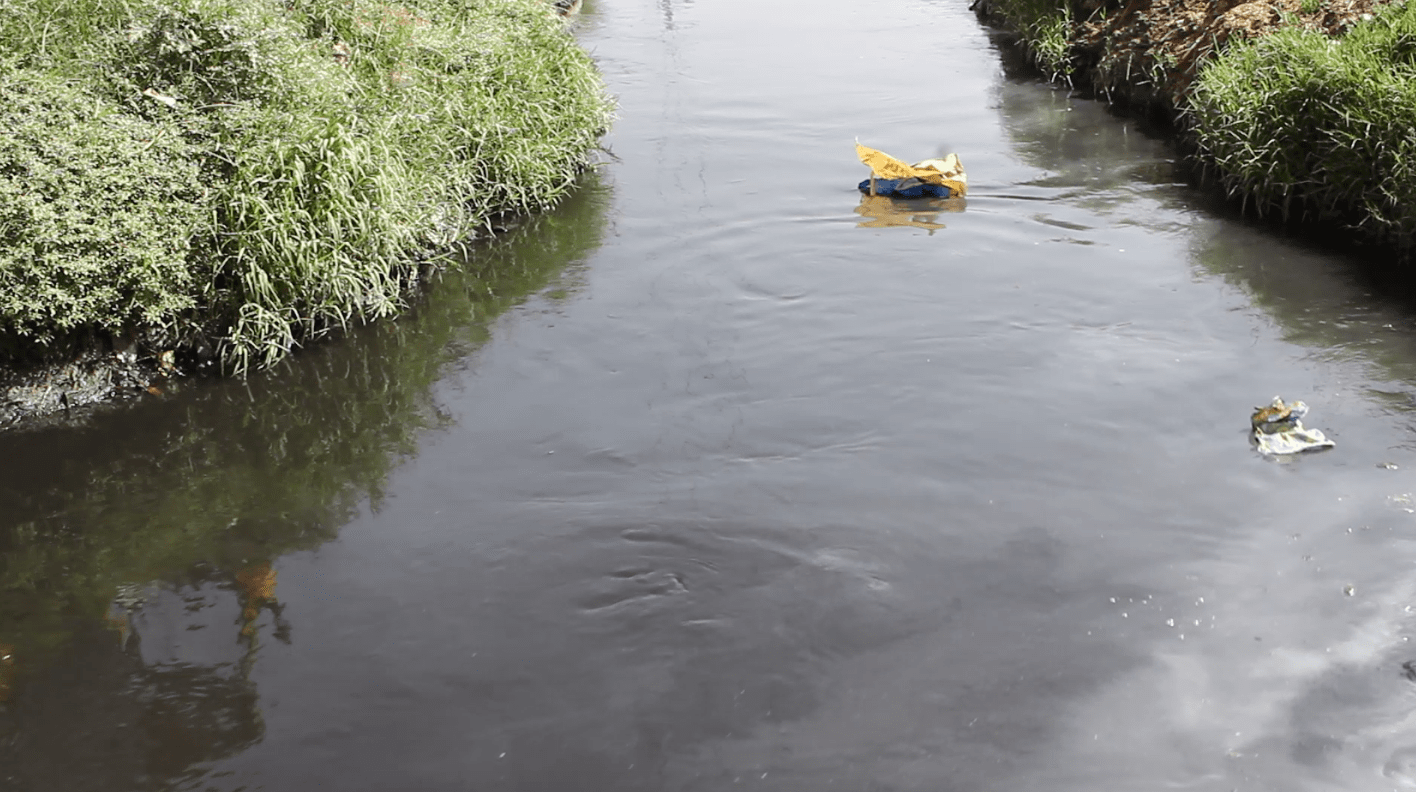} 
\includegraphics [width=3.5cm, height=2.3cm]{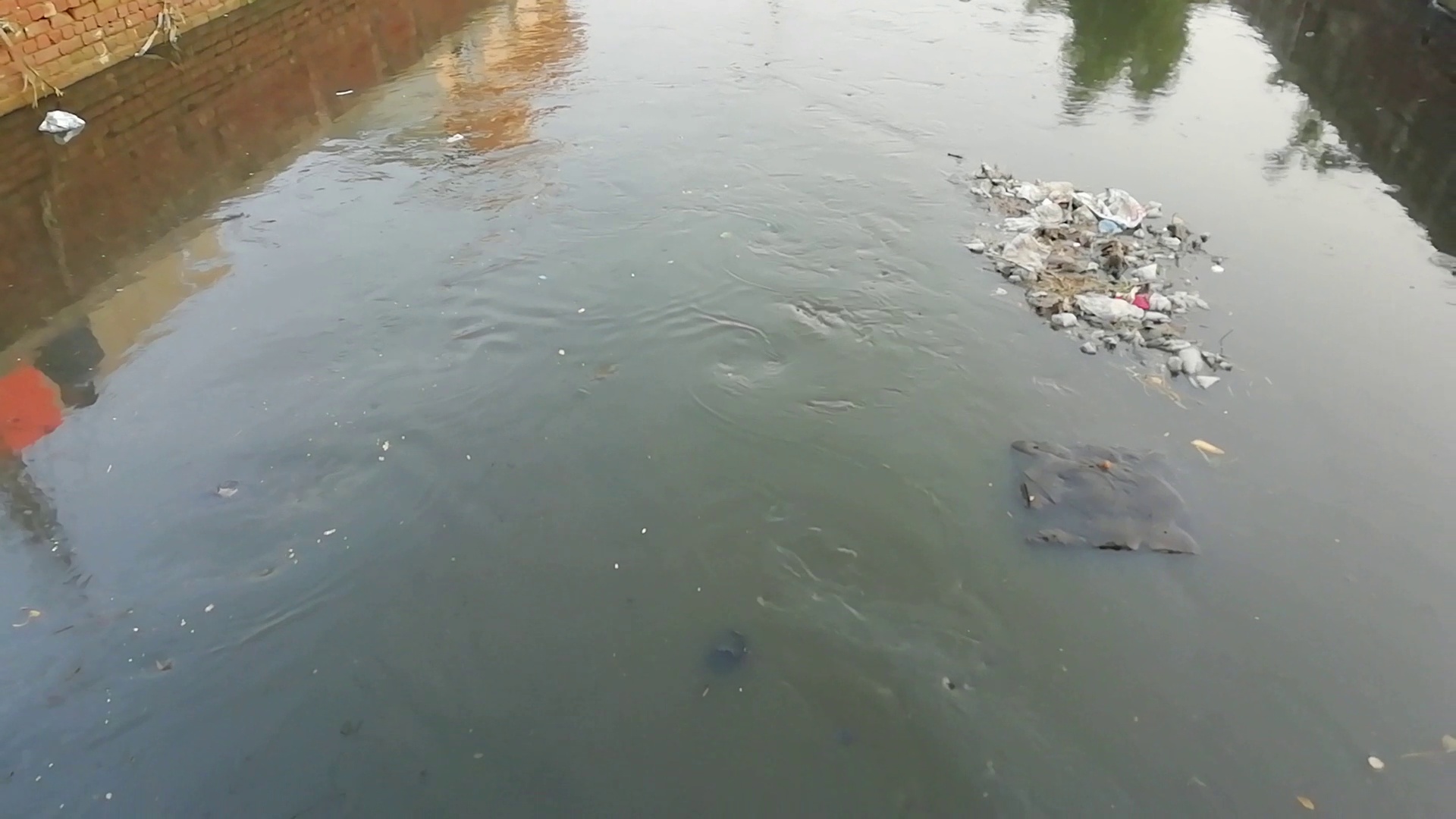}
\end{center}    
 \caption{Sample camera views from the collected dataset. The variation in views, shadows of overpass bridge, reflection of buildings and presence of vegetation is clear in these images.}
 \label{fig:dataset-samples}
    
\end{figure}

The existing work on vision-based approaches for detection of trash could be divided into three categories i) Classification of trash in a controlled environment, applicable at waste recycling plants~\cite{sakr2016comparing,sudha2016automatic}. ii) Detection of piles of trash, usually illegally dumped in cities~\cite{mittal2016spotgarbage,rad2017computer}. iii) Detection of sparse trash could be street trash or marine litter~\cite{fulton2018robotic,ge2016semi,valdenegro2016submerged,liu2018research}. In this paper, we introduce a fourth category of detecting visual trash floating on the water channels, especially drainage canals. Different from the above discussed studies, our problem focuses on surface trash present in canals running through dense urban areas.

Most of the recent work on trash detection employ deep learning based object detectors including SSD~\cite{liu2016ssd},  YOLO~\cite{redmon2018yolov3}, and Faster RCNN~\cite{ren2015faster}. These well knowm object detectors~\cite{liu2016ssd,redmon2018yolov3, ren2015faster,redmon2016you,wang2018pelee} are designed for general applications, especially for urban scenarios such as those related to surveillance and self driving cars. These networks do perform better on relevant benchmark datasets such as MS-COCO~\cite{chen2015microsoft} and Pascal-VOC~\cite{everingham2010pascal}. However, detecting trash over water channels is a more challenging problem due to the changes in object shape with flow of water and broad spectrum of object sizes. To overcome the issue of variation in object sizes, various efforts have been done such as image pyramids~\cite{Pang_2019}, feature fusion networks~\cite{fpn_2017,lin2017focal},  Thinned U-shape Modules(TUM)~\cite{M2Det2019aaai}, and attention mechanism~\cite{bello2019attention,vaswani2017attention}.

This paper introduces a new category of trash detection, provides a manually collected and annotated trash images dataset, and proposes a novel attention layer that implicitly focuses on smaller objects. Through experiment, we demonstrate that our proposed attention layer improves the detection of small trash particles missed by state-of-the-art object detectors~\cite{liu2016ssd,redmon2018yolov3,wang2018pelee}.


\begin{figure}[!t]
\centering
\captionsetup[subfloat]{font=footnotesize, justification=centering}
\begin{tabular}{cccc}
\begin{subfigure}{0.13\textwidth}
  \centering
   \includegraphics [width=2.3cm, height=2cm]{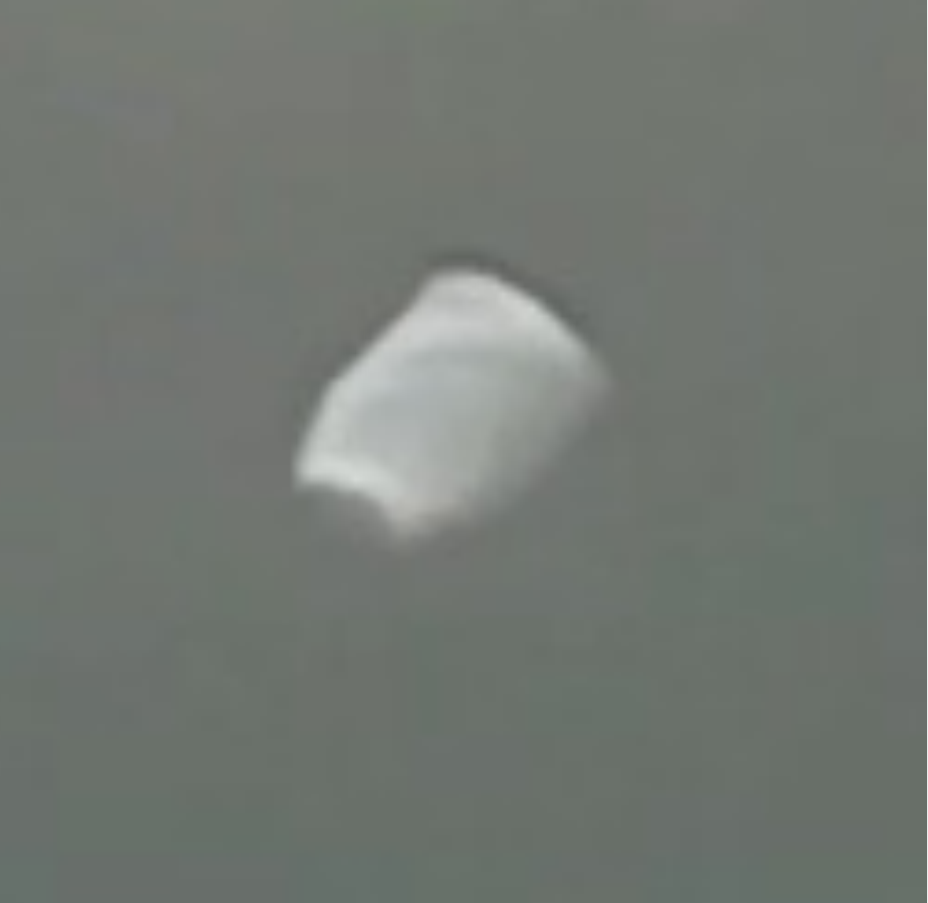} 
  \caption{Deformed}
  \label{fig:dormated_cup}
\end{subfigure}

\begin{subfigure}{0.13\textwidth}
  \centering
\includegraphics [width=2.3cm, height=2cm]{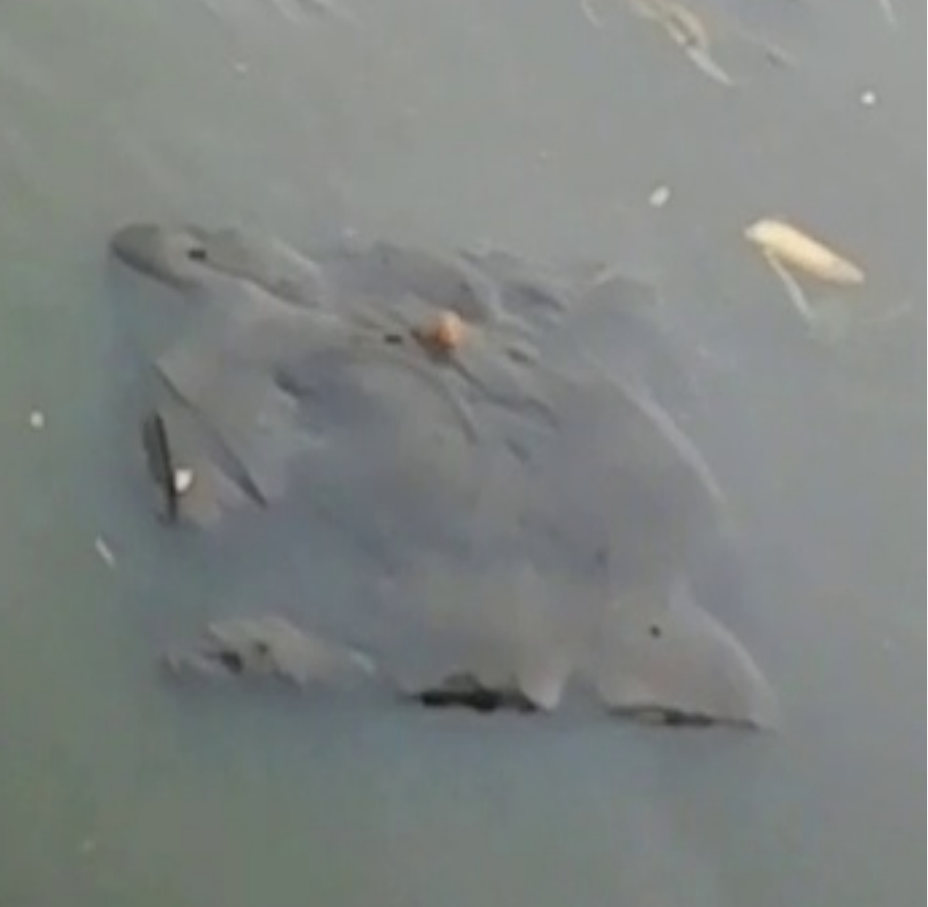} 
  \caption{Sub-merged}
  \label{fig:submerged}
\end{subfigure}

\begin{subfigure}{0.13\textwidth}
  \centering
   \includegraphics [width=2.3cm, height=2cm]{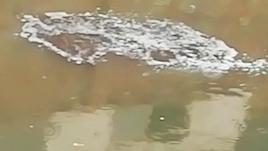} 
  \caption{Air bubbles}
  \label{fig:reflected_bird}
\end{subfigure} \\

\begin{subfigure}{0.13\textwidth}
  \centering
  \includegraphics [width=2.3cm, height=2cm]{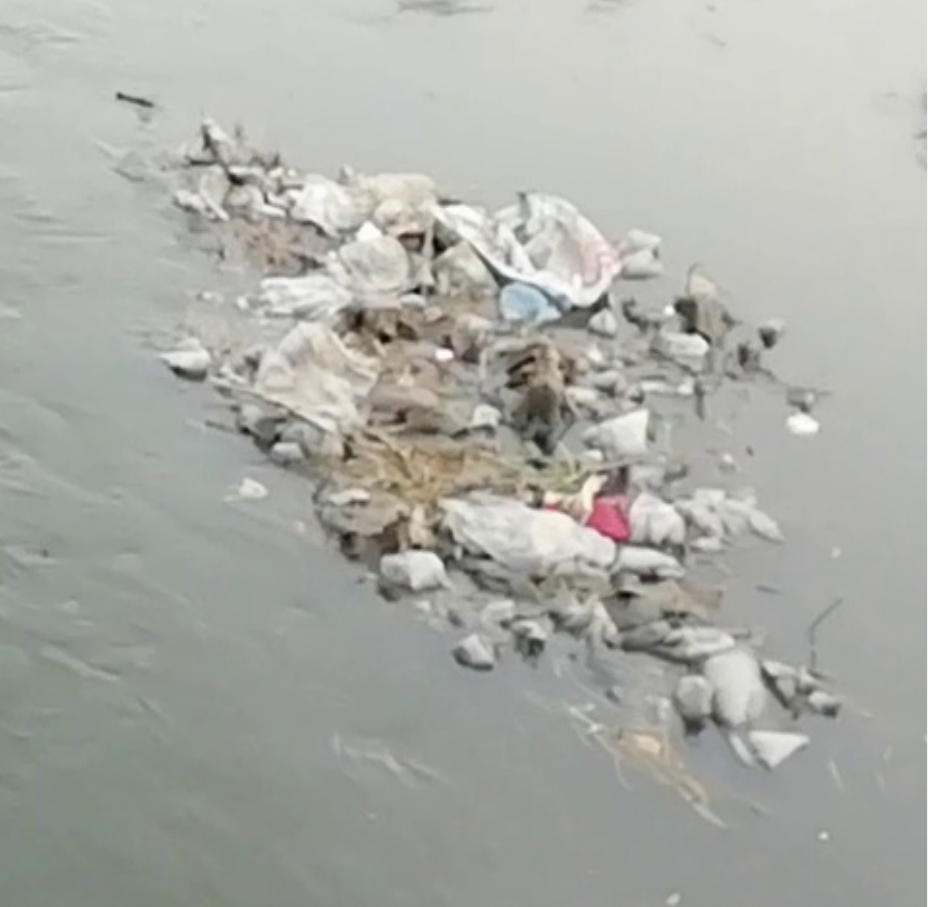}
  \caption{Piles of trash}
  \label{fig:Piles_1}
\end{subfigure}
\begin{subfigure}{0.13\textwidth}
  \centering
   \includegraphics [width=2.3cm, height=2cm]{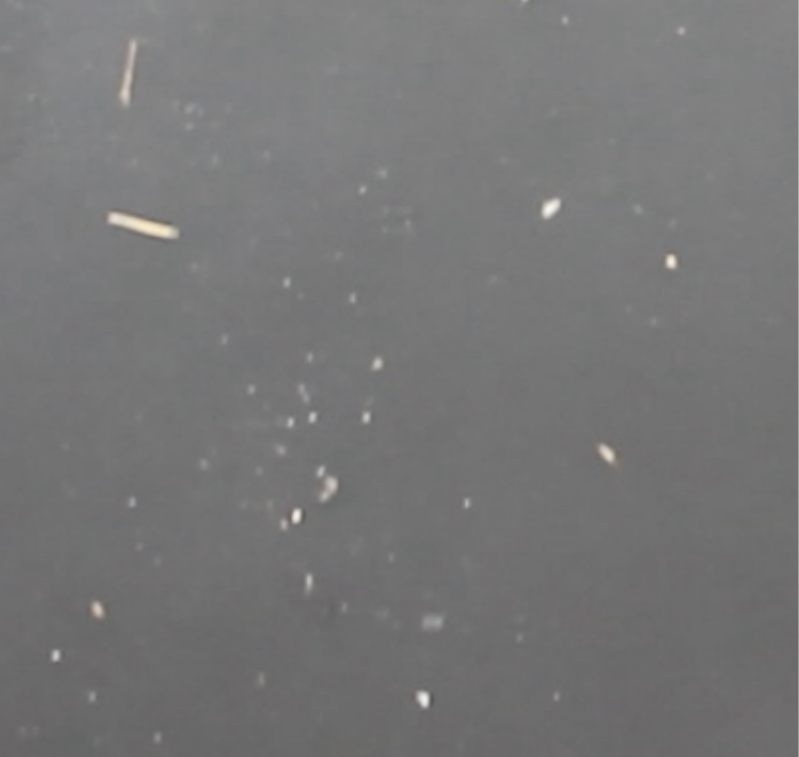}  
  \caption{Micro-particles}
  \label{fig:SparseTrash}
\end{subfigure}
\begin{subfigure}{0.13\textwidth}
  \centering
  \includegraphics [width=2.3cm, height=2cm]{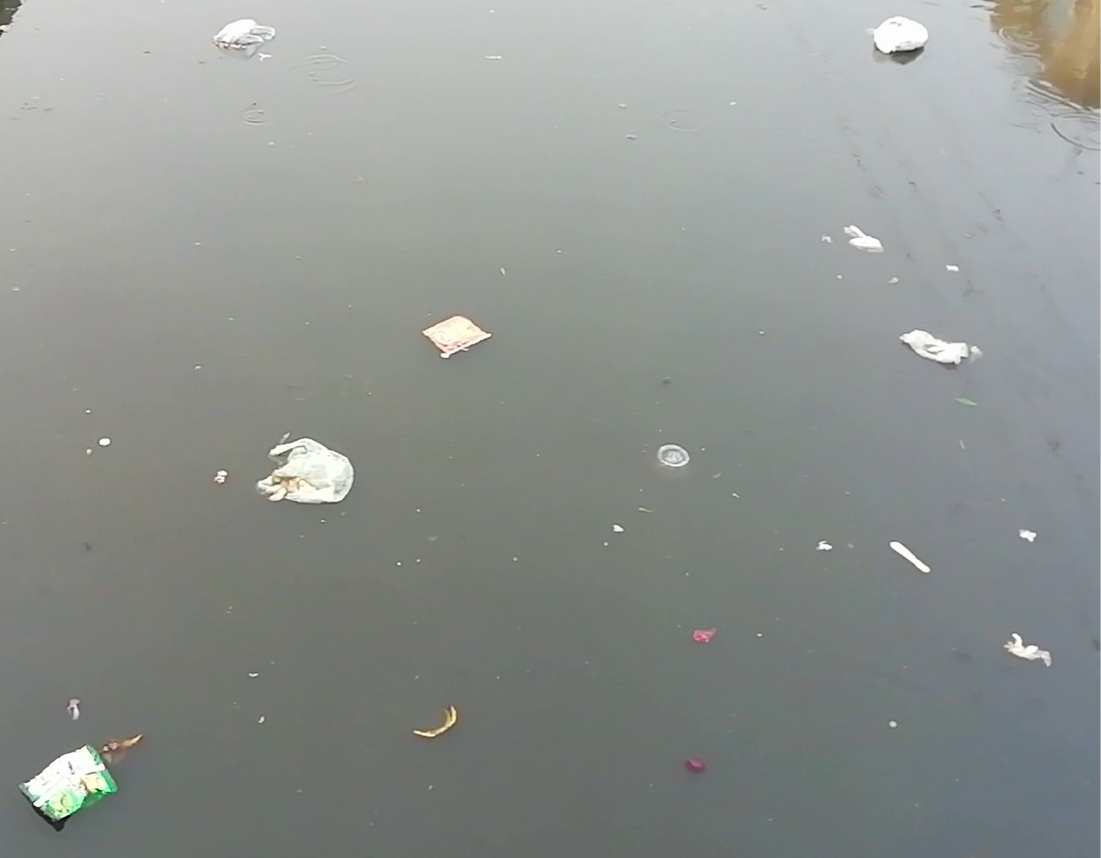}
  \caption{Sparse trash}
  \label{fig:bubble_large}
\end{subfigure} \\

\begin{subfigure}{0.13\textwidth}
  \centering
   \includegraphics [width=2.3cm, height=2cm]{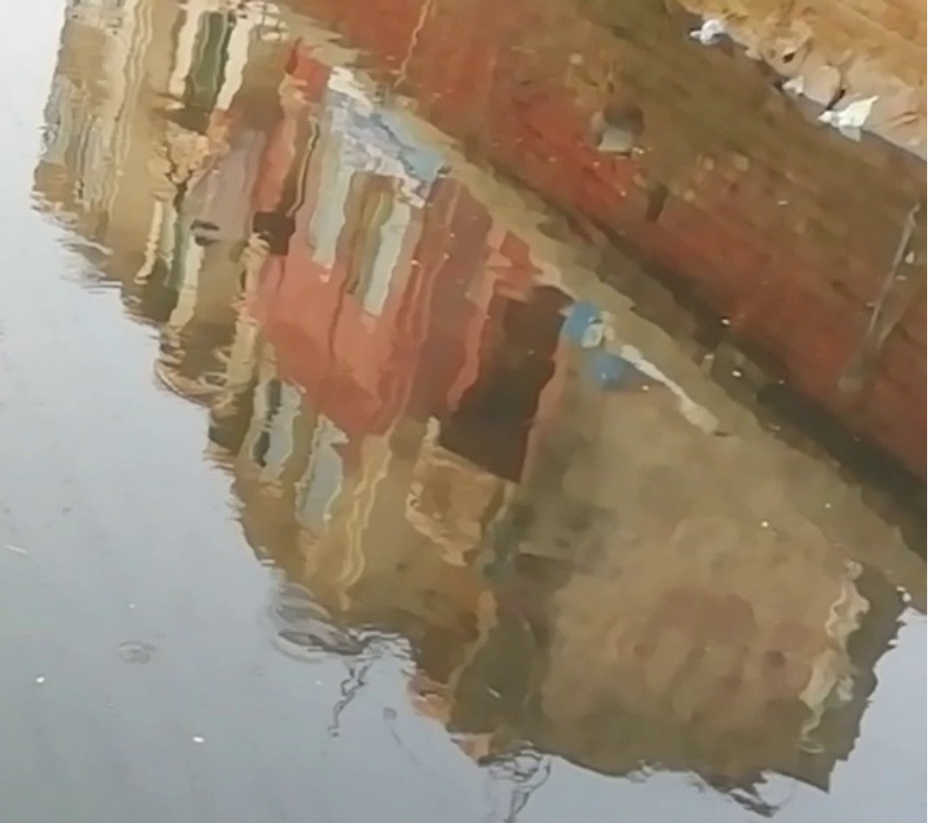} 
  \caption{Reflection of buildings}
  \label{fig:reflection_building}
\end{subfigure}
\begin{subfigure}{0.13\textwidth}
  \centering
    \includegraphics [width=2.3cm, height=2cm]{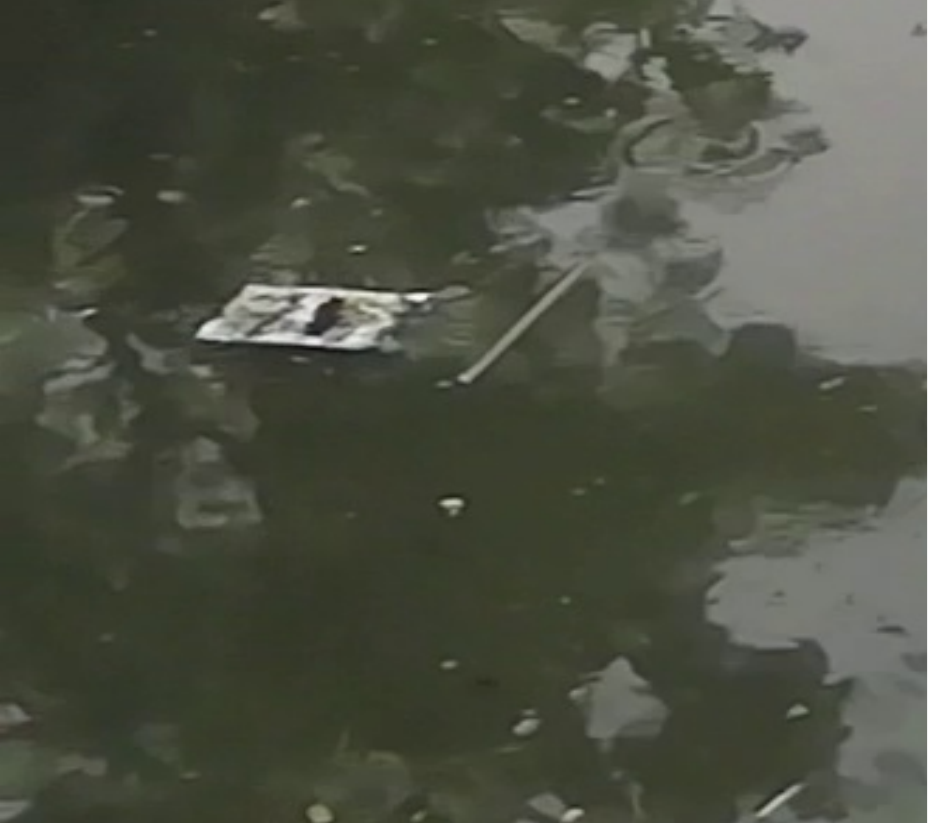}
  \caption{Object in Reflection}
  \label{fig:Reflection_1}
\end{subfigure} 
\begin{subfigure}{0.13\textwidth}
  \centering
  \includegraphics [width=2.3cm, height=2cm]{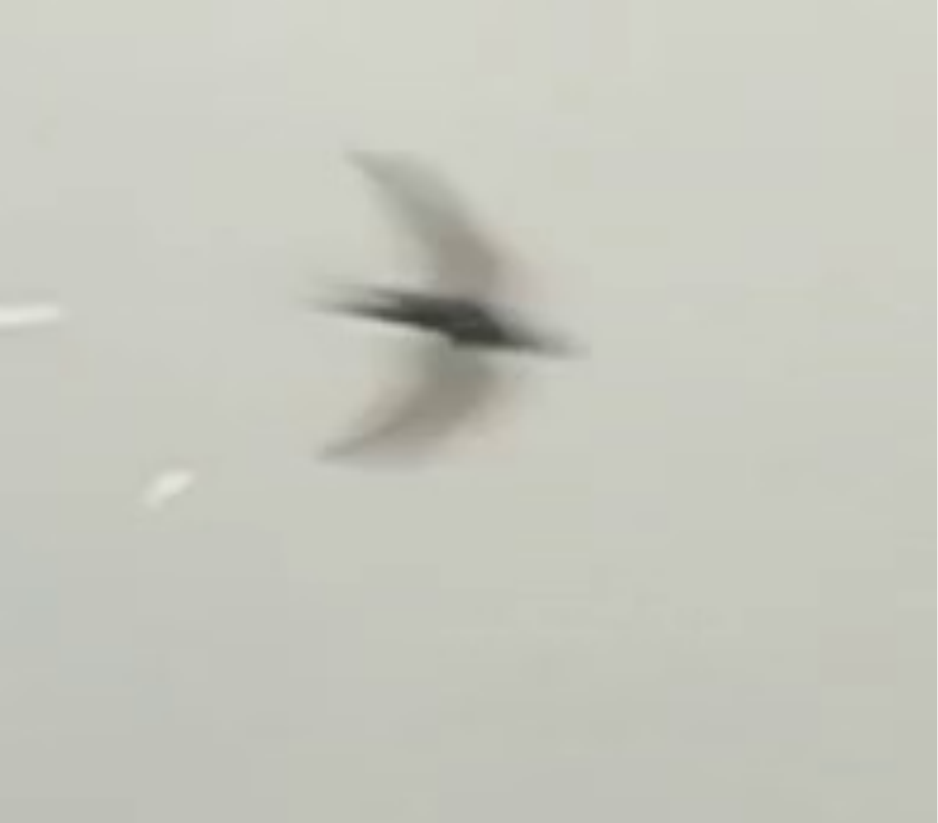}  
  \caption{Reflection of flying bird}
  \label{fig:MicroTrash_1}
\end{subfigure} 
  \end{tabular}
 \caption{Sample cropped images showing wide-variety of challenging scenarios present in the fresh water and drainage canals. 
 }
 \label{fig:dataset-challenges}
    
\end{figure}

\section{Dataset} 
\label{sect:dataset}
\subsection{Collection}
\label{sect:dataset-collection}
Although, the problem of ocean trash has received significant attention in the recent years, however, to the best of our knowledge the problem of trash in fresh and waste water ways has not been addressed in the past. Consequently, there is no any existing dataset available on this problem. Thus, in this work, we contribute first of its kind image dataset. The videos for dataset were collected during different day times, weather conditions, and localities to ensure the recorded data contain a myriad of objects of interest. We surveyed many sites near commercial areas, slum neighbourhoods and industrial areas of the city and selected five critical sites. We recorded $30$ different videos of upto $60$ minutes each. Some of the example images from our dataset are shown in Fig.~\ref{fig:dataset-samples}. These images contains several challenging scenarios which are discussed next.

\subsection{Challenges in dataset}
\label{subsect:challeges}

Objects in the water are often deformed (see Fig.~\ref{fig:dataset-challenges}(a)), have no defined geometrical shape and their shapes also change over time. For example, floating plastic bags may distort to a multitude of shapes that vary with time. Due to water flow, not only objects are submerged (Fig.~\ref{fig:dataset-challenges}(b)) but also sometimes sink in water and then resurface later. Sometimes sewer gases are produced through the decomposition of organic household products or industrial waste as shown in Fig.~\ref{fig:dataset-challenges}(c), and they show resemblance to surface trash. As shown in Fig.~\ref{fig:dataset-challenges}(d)-(f), the trash may be present as sparsely distributed objects or dense piles of trash on the water surface. 

Since the dataset was collected from dense urban areas, the reflections and shadows of static (buildings and electricity poles, and moving (flying birds) objects (see Fig.~\ref{fig:dataset-challenges} (g)-(i)) appear significantly in these water channel. Depending upon the camera view point and time of the day, they may cover a significant portion of the water surface. The color and opacity of water varies from channel to channel depending on the amount of chemical discharge from factories and sewerage, eventually changing the reflection and refraction of sunlight. These cases are not found in ocean environment and increases the complexity of our problem.

\setlength{\tabcolsep}{0.1cm}
\begin{table}[t]
\begin{center}
\caption{Distribution of objects in annotated images when divided into small, medium and large categories.  }\label{Table:dataset_dist}
\vspace{0.03cm}
\begin{tabular}{|l|c|c|}
\hline
Size & No. of Objects & Area / px$^2$  \\ \hline
Small &  11090 & area $\leq$ $32^2$ \\
Medium &  33116 & $32^2$ $<$ area $\leq$ $96^2$ \\ 
Large &  4692  &  area $>$ $96^2$  \\
\hline
\end{tabular}
\end{center}

\end{table}
\setlength{\tabcolsep}{0.02cm}

\subsection{Annotation}
\label{subsect:annotation}
Annotations were done by four different individuals under the supervision of a domain expert. A total of $13500$ images were selected from the collected videos at a regular interval and annotated for bounding box in VOC format~\cite{everingham2010pascal}. LabelImg\footnote{https://github.com/tzutalin/labelImg} was used to annotate the images for bounding boxes. A total of $48898$ objects were annotated in $13500$ images ranging from almost $256$ to $300,000$ px$^2$ in area. From this dataset $12500$ images were considered as the train-validation set and the remaining $1000$ test images were divided into easy and hard test sets containing $500$ images each. Images in \emph{Easy Test Set} have different weather conditions and water texture and some videos were collected during rainy weather (No examples of rainy day in training set) whereas \emph{Hard Test Set} contains a variety of view points, different weather conditions, water color and gas bubbles coming out of water.

 We intentionally do not annotate micro-particles, leaves, twigs, non-floating trash and air bubbles. Minuscule micro-particles affect the texture of the water surface, so instead of object detection, they can be quantified using texture analysis and hence were omitted from our study. Moreover, we also did not annotate objects which were less than $7$ pixels in both width and height. To get the distribution of annotated objects in terms of size, following MS-COCO standard they were grouped into three categories i.e. small with area less than $32^2$ px$^2$, medium with area between $32^2$ to $96^2$ px$^2$ and large otherwise. Table~\ref{Table:dataset_dist} shows that the more than $90\%$ of objects are either small or medium size.

\begin{figure}[t]
    \centering
\begin{tabular}{cc}
\begin{subfigure}{0.3\linewidth}
 \centering
  \includegraphics[height=3cm]{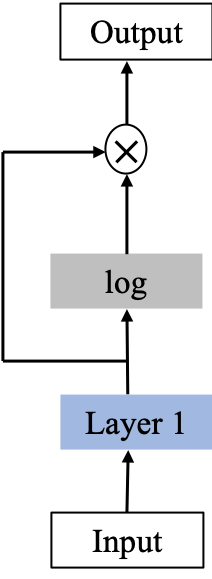}
  \caption{Attention layer}
 \label{fig:qualti-pelee}
 \end{subfigure} 
\begin{subfigure}{0.6\linewidth}
 \centering
  \includegraphics[height=3cm]{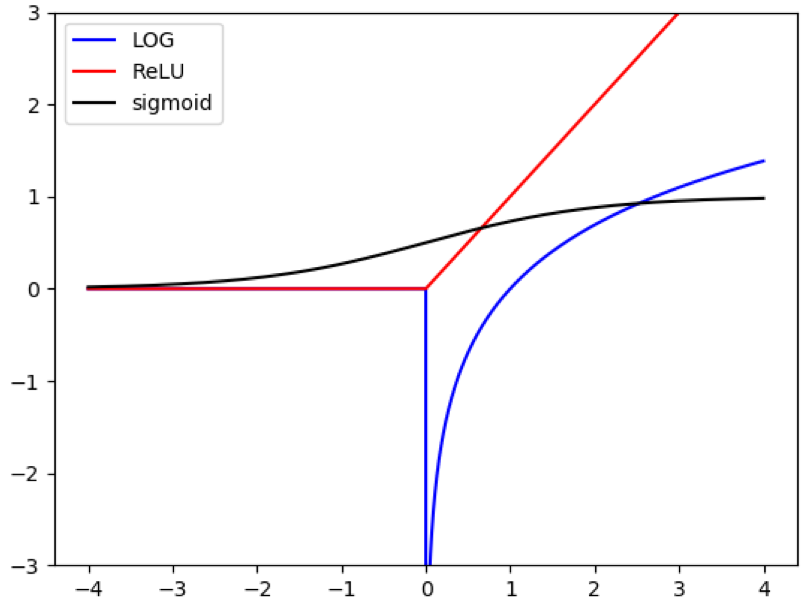}
  \caption{Activation Functions}
 \label{fig:qualti-pelee2}
 \end{subfigure} 
 \end{tabular}
 
\caption{Attention Layer multiplies output of a convolutional layer with its log activation.}
\label{fig:attn_layer}
\end{figure}

\section{Proposed Method}

\label{proposed_method}

We trained some of the popular object detectors such as SSD~\cite{liu2016ssd} and YOLO-v3-Tiny whereas YOLO-v3~\cite{redmon2016you} and PeeleNet~\cite{wang2018pelee}. We observed that these models have encouraging results on our dataset as given in Table~\ref{Table:results2}. 
Further analysis (see Table~\ref{Table:results3}) indicates that the performing models shown unsatisfactory outcome on tiny objects. The feature extractors used in these object detectors merge features from multiple scales to detect objects of variable sizes. Despite this, they still fail to identify smaller objects present in our novel case of visual trash detection. To resolve this problem, an attention layer can be employed to enforce the algorithm to focus on smaller objects.

Conventionally, attention layer employ \emph{Sigmoid} or \emph{Softmax} to predict the probability of objectness in features, which are then merged together to highlight a certain area in the feature space. Attention based on \emph{Sigmoid} and \emph{Softmax} activation functions seems to perform well for segmenting the pixels near boundaries of the objects~\cite{}. Nevertheless, in our study, we require an activation function that implicitly focuses on smaller objects in an image. \emph{Logarithm} scale is used to overcome the skewness in the data i.e. if few values are very large or very small than rest of the data, then logarithm would transform this wide-range into a smaller one. Thus, it reduces the variance in the features and scales up smaller values. This motivates us to utilize \emph{log} based attention layer to emphasize on smaller objects. We employed \emph{log} attention layer as given in Fig.~\ref{fig:attn_layer} on multi-resolution features. Mathematically, we define our attention layer as:
 \begin{equation}
 \label{log_attention}
    f_{i+1} = f_i \times log (ReLU(f_i) + 1),
 \end{equation}
where $f_i$ is output of $i^{th}$ layer, and $i = {0,...,N}$ and $N$ is number of layers. Here, \emph{ReLU} discards the negative values in the activations and bias $1$ shifts it one scale up, making it possible to compute \emph{log}. The derivative of this attention layer would be: 

\begin{equation}
 \label{log_attention2}
    \bigtriangledown f_{i+1} = 
    \begin{cases}
     log (f_i + 1) + \frac{1}{f_i + 1} & f_i\ge 0 \\
    1 & otherwise
    \end{cases}
 \end{equation}

\begin{figure}[t]
\centering
 \includegraphics[width=0.9\columnwidth]{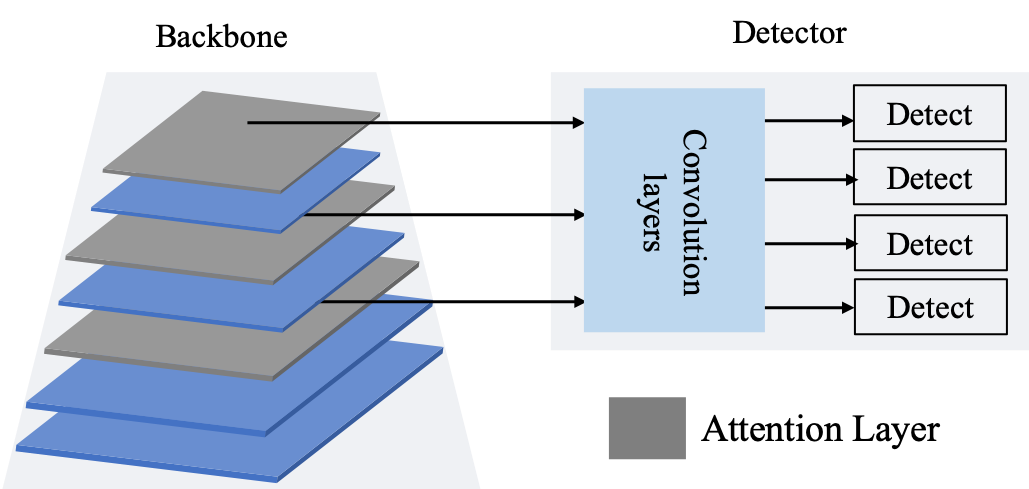}
\caption{Proposed model: Left side backbone network is modified to introduce attention layer while the right side is the detector.}
\label{fig:model}
\end{figure}

\begin{figure}[h]
\centering
\captionsetup[subfloat]{font=footnotesize, justification=centering}
\begin{tabular}{ccc}

\begin{subfigure}{0.14\textwidth}
  \centering
   \includegraphics [width=2.34cm, height=2.27cm]{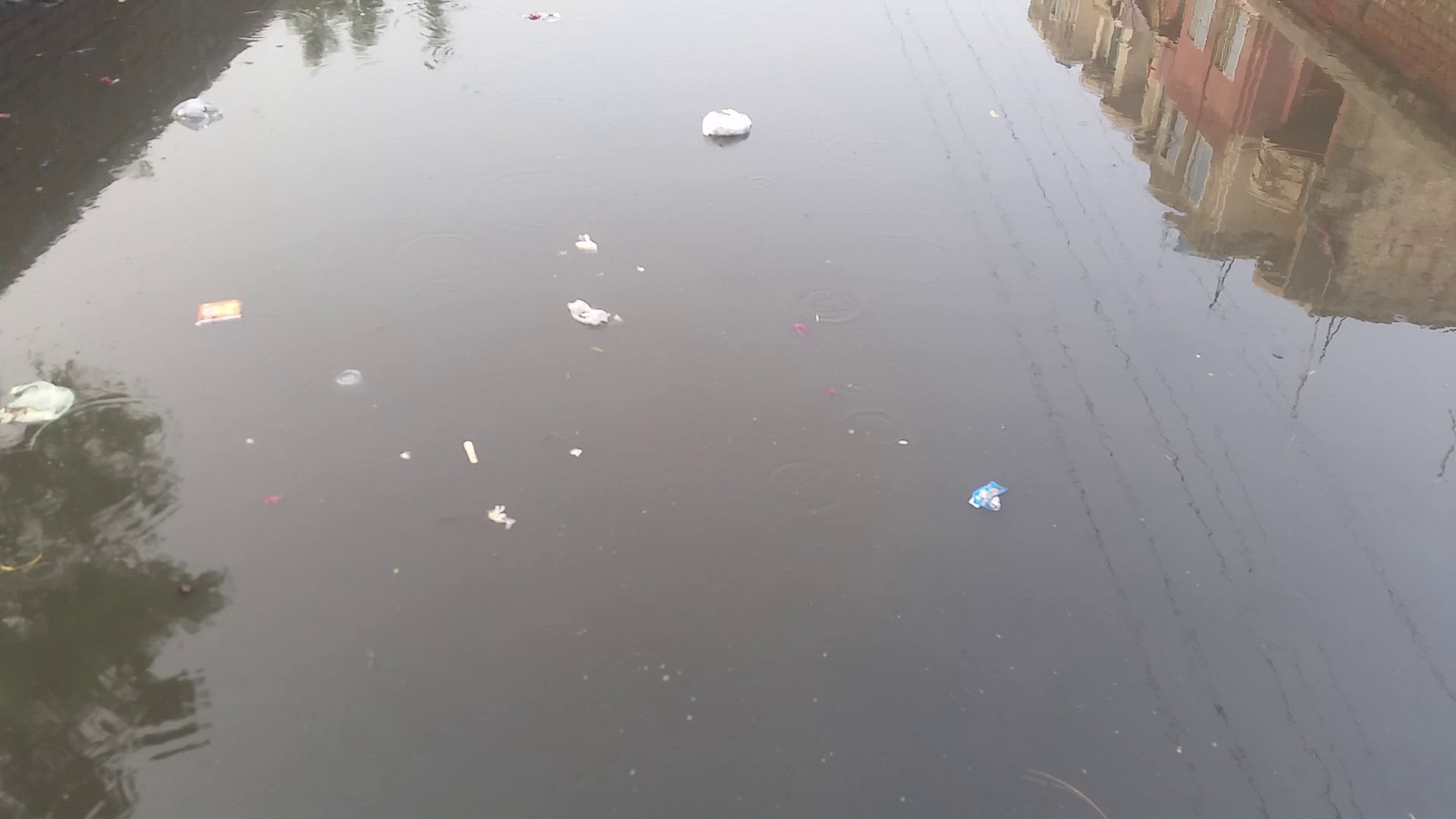} 
  \caption{}
  \label{fig:input}
\end{subfigure}
\begin{subfigure}{0.14\textwidth}
  \centering
    \includegraphics [width=2.35cm, height=2.3cm]{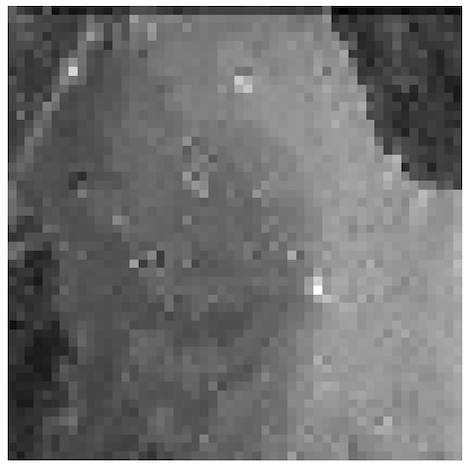}
  \caption{}
  \label{fig:act_vanilla}
\end{subfigure} 
\begin{subfigure}{0.14\textwidth}
  \centering
  \includegraphics [width=2.35cm, height=2.3cm]{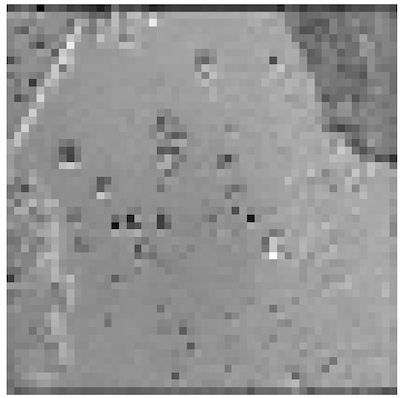}  
  \caption{}
  \label{fig:act_log}
\end{subfigure} 
  \end{tabular}
 \caption{Visualization of layer activation: Effects of log attention layer on features learned by YOLO-v3 same convolutional layer. (a)Input Image (b) Vanilla YOLO-v3 (c) Log attention.
 }
 \label{fig:activation}
    
\end{figure}

The \emph{log} attention introduces numerical stability by responding to unevenness in the features due to large variations in object size. Object detector contains a backbone network that learns deep features, some optional convolutional layers applied on features learned by backbone and final prediction layers. Since, the backbone network learns the features required by the detector, so amplification of smaller activation values in backbone network would eventually drive the performance of detector. Therefore, we applied our attention layer on the features consumed by the detector as shown in Fig.~\ref{fig:model}. Updated features were forwarded to preceding layer of the backbone network to improve features progressively as evident in Fig.~\ref{fig:activation}, where activation show clear improvement learning features for objects.  

\section{Results and Evaluation}
\label{sect:results}
We compare the performance of our model with state-of-the-art models such as SSD, YOLO-v3, YOLO-v3-Tiny, PeeleNet. Since there is no existing dataset on the problem of trash on water surface, so these methods were evaluated on the dataset introduced in this paper (Section~\ref{sect:dataset}). In order to validate the performance of the trained network, we used two standard benchmark performance metrics namely average precision (AP) and Intersection over Union (IoU) as used by MS-COCO~\cite{chen2015microsoft}.

\subsection{Training}
\label{subsect:training}
All the networks were trained using their default hyper-parameters and APIs. 
All the models were initialized with pre-trained weights trained on Pascal VOC~\cite{everingham2010pascal}. The $12500$ images of the dataset were randomly split into $80-20\%$ and these splits were fixed for all the models. We used $80\%$ of them for training, $20\%$ for validation during training. The test set was made from the images of site other than the train-validation sites. It contains a total of $1000$ images which were sub-divided into two sets i.e Easy and Hard Test Set, containing 500 images each. 

\setlength{\tabcolsep}{0.01cm}
\begin{table}[t]

\caption{Comparative evaluation of the state-of-the-art object detection techniques. (Key: AP: Average Precision, IoU: Intersection over Union).}

\label{Table:results2}
  \centering
  \begin{adjustbox}{width=0.97\linewidth}
    \begin{tabular}{|l|c|c|c|c|}
    \hline
    \multicolumn{1}{|c|}{\bfseries Model} & \multicolumn{2}{|c|}{\bfseries Easy Test Set} &
    \multicolumn{2}{|c|}{\bfseries Hard Test Set} \\ 
    &AP&IoU & AP&IoU \\ \hline
    SSD~\cite{liu2016ssd}  &24.1 & 64.0 & 26.3 & 72.0 \\
    YOLO-v3-Tiny  &5.6 & \textbf{69.2} & 11.6 & 66.9 \\
    YOLO-v3~\cite{redmon2018yolov3}  &43.8 & 64.5 & 31.5 & 68.5 \\
    
    RetinaNet~\cite{lin2017focal} &45.6 & 73.7 & 41.0 & 74.0 \\

    RetinaNet-resnet50 &49.9 & - & 48.5 & - \\
    
    PeleeNet~\cite{wang2018pelee} &40.7 & 67.1 & 24.2 & 72.1\\
    \hline
    YOLO-v3+Attn  &\textbf{48.1} & 64.5 & 31.2 & 69.4 \\
    PeleeNet+Attn & 41.4 & 66.4 & 23.5 & \textbf{72.7} \\
    RetinaNet+Attm &\textbf{51.8} & 73.7 & 43.9 & 73.9\\ RetinaNet-resnet50+Attn & 52.6 & - & 43.9 & - \\
    \hline
    \end{tabular}
     
   \end{adjustbox}
 \end{table}
 \setlength{\tabcolsep}{0.03cm}

\subsection{Quantitative Results}
\label{subsect:quant-results}

The quantitative evaluation of the object detection algorithms on two sets (easy and hard) is shown in Table~\ref{Table:results2}. Our proposed attention layer with YOLO-v3 outperforms all the other models on the Easy Test Set whereas it approximately gives the same performance on the Hard Test Set. SSD and YOLO-v3-Tiny fail to learn and given poor average precision for both test sets. 

 All models have comparable IoU but in terms of average precision (AP), YOLO-v3 with our attention layer outperforms all other models on the Easy Test Set with an AP score of $48.1\%$ whereas it closely coincides with the Vanilla YOLO-v3 on the Hard Test Set. YOLO-v3-Tiny has the lowest AP score of $11.6\%$. 

\begin{table}[t]
\begin{center}
    
\caption{Comparative analysis of average precision (AP) scores of the state-of-the-art object detection techniques on three different object size categories namely small(S), medium(M) and large(L) indicated by superscript.}
  \centering
  \begin{adjustbox}{width=0.97\linewidth}

    \begin{tabular}{|l|c|c|c|c|c|c|c|c|c|}
    \hline
    \multicolumn{1}{|c|}{\bfseries Model}& \multicolumn{3}{|c|}{\bfseries Easy Test Set} &
    \multicolumn{3}{|c|}{\bfseries Hard Test Set}\\ 
    & $AP^S$& $AP^M$ & $AP^L$ & $AP^S$& $AP^M$ & $AP^L$ \\ \hline
    SSD~\cite{liu2016ssd}& 1.6 & 7.8 & 38.1 & 0.8 & 10.2 & 31.3 \\
    YOLO-v3-Tiny & 0.0 & 1.9 & 36.7 & 0.6 & 3.1 & 12.4 \\
    YOLO-v3~\cite{redmon2018yolov3} & 4.5 & 16.3 & 52.0 & 1.3 & 9.9 & 32.2 \\

    RetinaNet~\cite{lin2017focal}& 5.0 & 28.6  & 71.8 & 4.0 & 19.5 & 38.9 \\
    
    RetinaNet-resnet50 & 3.7 & \textbf{36.4} & 67.6 & 5.0 & 24.5 & 40.5 \\
    PeleeNet~\cite{wang2018pelee} & 5.5 & 15.3 & 50.3 & 1.9 & 9.4 & 28.1\\
    \hline
    YOLO-v3+Attn & 5.4 & 16.3 & 51.6 & 1.3 & 10.0 & 31.8 \\
    
    PeleeNet+Attn & 6.2 & 15.1 & 46.6 & \textbf{9.0} & 8.7 & 30.1\\
    
    RetinaNet+Attn& \textbf{6.3} & 35.8  & \textbf{75.5} & 4.1 & \textbf{22.2} & \textbf{41.5} \\
    
    \hline
    \end{tabular}
    \end{adjustbox}
    \label{Table:results3}
\end{center}
\end{table}

\subsection{Analysis on object sizes}
\label{additional-analysis}

In order to find out the performance dependency on object sizes, the trained networks were evaluated on three scales of objects given in Table~\ref{Table:dataset_dist}. Table~\ref{Table:results3} demonstrates that the performance of all the networks on smaller objects is poorer than medium and large objects. Even though large objects are only $4692$, which is  half of the number of smaller objects, networks were still able to detect them. This is due to prior training of models for object detection task on Pascal VOC dataset, so the information of 'objectness' was retained for large objects. YOLO-v3 with attention has better AP on large and medium object on the \emph{Easy Test Set} and performs better on large objects on the \emph{Hard Test Set}. PeeleNet with attention closely coincides with YOLO-v3 with attention for all object sizes.


\section{Conclusion}
\label{conclusion}
This paper presents a new category of visual trash detection through deep learning based object detectors. A dataset of trash floating on canal surface in dense urban areas is collected and annotated. Then, several recent and popular deep object detection models were trained and evaluated. Finally, we proposed a novel \emph{log} based attention layer that has improved the performance particularly on small objects. 
Overall, the detection of floating trash specially in water channels in urban areas is a challenging task and an emerging area of research. The dataset provided in this work will serve as a stepping stone towards finding a solution to this problem.







\bibliographystyle{IEEEtran}
\bibliography{main.bib}



\end{document}